\documentclass{article}
\usepackage{spconf,amsmath,graphicx}

\usepackage{cite}
\usepackage{amsmath,amssymb,amsfonts}
\usepackage{algorithmic}
\usepackage{graphicx}
\usepackage{textcomp}
\usepackage{xcolor}
\usepackage[font=scriptsize]{subcaption}

\usepackage{url}
\usepackage{multirow,booktabs}
\usepackage{breqn}
\usepackage{rotating, makecell}

\title{Cascaded Generation of High-quality Color Visible Face Images from Thermal Captures}
\name{Naser Damer$^{12}$, Fadi Boutros$^{12}$, Khawla Mallat$^{3}$, Florian Kirchbuchner$^{12}$, Jean-Luc Dugelay$^{3}$, Arjan Kuijper$^{12}$}
\address{$^{1}$Mathematical and Applied Visual Computing, TU Darmstadt,
Darmstadt, Germany\\ $^{2}$Fraunhofer Institute for Computer Graphics Research IGD,
Darmstadt, Germany\\ $^{3}$Digital Security Department, EURECOM, Sophia Antipolis, France}

\begin{document}
%
\maketitle
\begin{abstract}
Generating visible-like face images from thermal images is essential to perform manual and automatic cross-spectrum face recognition. 
We successfully propose a solution based on cascaded refinement network that, unlike previous works, produces high quality generated color images without the need for face alignment, large databases, data augmentation, polarimetric sensors, computationally-intense training, or unrealistic restriction on the generated resolution. 
The training of our solution is based on the contextual loss, making it inherently scale (face area) and rotation invariant. 
We present generated image samples of unknown individuals under different poses and occlusion conditions.
We also prove the high similarity in image quality between ground-truth images and generated ones by comparing seven quality metrics. We compare our results with two state-of-the-art approaches proving the superiority of our proposed approach.
\end{abstract}
\vspace{-5mm}
\section{Introduction}
\vspace{-3mm}

The deep learning-driven surge in face recognition accuracy enabled and widened the deployment of face recognition in challenging scenarios. Such scenarios may not benefit from subject collaboration or optimal capture environment, such as suboptimal illumination. This was a major drive behind face recognition in the thermal spectrum.
Interest in thermal imaging technology has significantly grown recently, and thermal cameras have evolved to become affordable and integratable. However, they still suffer low performances due to low image resolution, lack of textural and geometrical information, and shortage in available face training data. Therefore, generating high-quality face images from thermal captures is essential to enable automatic and manual face comparisons. 





Previous works faced challenges in producing high quality visible images from thermal images. Using more informative, yet less cost efficient and less integrable, polarimetric thermal images enhanced the quality of the generated images to some degree \cite{HZhang17,Riggan18}. Most of these previous works were limited to generating gray-scale visible images and did not consider occlusion. Recent works proposed using generative adversarial networks (GAN) to make the conversion \cite{HZhang17,TZhang18}. However, the computationally-needy, data-hungry, low-resolution nature of GANs lead to less than optimal results given the limited data availability and the possible negative effect of data augmentation on an accurate conversion.

In this work, we successfully propose the use of cascaded refinement networks coupled with the contextual loss to generate high quality colored visible images from thermal captures. This eliminated the need of computationally-intense training and large aligned databases. Moreover, our solution is inherently rotation invariant and does not require source image alignment. 
To prove the success of our solution, we present visible samples of generated images in different poses and occlusion scenarios. 
A quantitative analysis of generated image quality is performed proving the high similar quality properties of the generated image to the ground-truth images. We compare our results with two recently proposed approaches proving higher image properties similarity to the ground-truth and higher visual quality of our solution. 

\vspace{-3mm}
\section{Related work}
\label{sec:rw}
\vspace{-2.5mm}

Cross spectrum face recognition can be seen under two categories, cross spectrum matching and cross spectrum generation. The matching aims at direct comparison of images captured under different spectrum. The generation category brings images to a common spectrum to enable both, manual face comparison by experts and automatic face recognition. 

Although some works adopting the cross spectrum matching approach have reported significant performances \cite{PLS15,Sarfraz15}, it is challenging for these techniques to be integrated with existing commercial and academic face recognition systems. Moreover, further development of such techniques requires large cross spectrum databases
. Such a solution eliminate the possibility of manual comparison required for some legal processes. Therefore, in this work, we focus on generating a visible-like face image from thermal captures.




Li et al. \cite{Li07} were one of the earliest researchers to investigate thermal to visible face generation. They proposed learning-based framework that exploits the local linearity in the spatial domain of the image and also in the image manifolds followed by applying Markov random field to improve the hallucinated visible face. Considering  the recent breakthroughs in deep learning, Zhang et al. \cite{HZhang17} employed deep generative models, particularly Conditional Generative Adversarial Neural network to synthesize visible faces from polarimetric inputs. Riggan et al. \cite{Riggan18} proposed a convolution neural network that aims to map edge-based features both locally and globally from the polarimetric to the visible domain. Even though these two solutions were evaluated on conventional thermal data, they are designed for polarimetric data that capture more geometric and textural information. Moreover, the above mentioned works generated visible face images in gray scale and discarded face generation under operational conditions such as head pose and occlusion variations.
A recent work by Zhang et al. \cite{TZhang18} considered synthesizing colored faces from thermal images with various head poses and occlusion with eyeglasses. However, it is still very different from our framework as they used  Conditional Generative Adversarial Neural network. Given that GANs are data hungry, they had to use data augmentation to train the model, resulting in generated images with high level of visible artifacts. Moreover, the work used aligned images and did not mention the generated images resolution, which is commonly limited by the computational and data requirements of the GAN.  


%
%
%
%
%

\vspace{-3mm}

\section{Methodology}
\label{sec:meth}
\vspace{-3mm}

\begin{figure}[h]
\centering
\includegraphics[width=1.0\columnwidth]{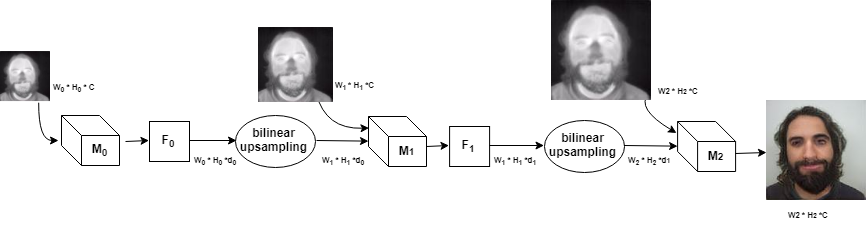}
\vspace{-6mm}
\caption{The CRN-based multi-scale approach to transform thermal images into visible-like images. Here, only three consecutive modules are shown as an example. These cascaded modules can be repeated to reach the targeted resolution.  }
\label{fig:CRN}
\vspace{-4mm}
\end{figure}

\textbf{Proposed approach:} To generate images, we propose to base our approach on the cascaded refinement network (CRN) \cite{DBLP:conf/iccv/ChenK17}. We chose the CRN as the basic block for our image generation as it considers multi-scale information and based on training a limited number of parameters. This allows for a higher resolution generation and less data size dependency in comparison to solutions based on GAN. CRN is a convolutional neural network that consists of inter-connected refinement modules. Each module consists of only three layers, input, intermediate, and output layer. The first module considers the lowest resolution space (4x4 in our case). This resolution is duplicated in the successor modules until the last module (128x128 in our case), matching the target image resolution. For more detailed description of the CRN, one can refer to \cite{DBLP:conf/iccv/ChenK17}.

To control the training of our CRN network, we used the contextual loss function (CL) \cite{DBLP:conf/eccv/MechrezTZ18}. This choice is based on our need for: a) a loss function that is robust to not well aligned data (as in our use-case where input face images are not uniformly aligned), and b) neglect outliers on the pixel level (in comparison to pixel level loss \cite{DBLP:conf/cvpr/IsolaZZE17,DBLP:conf/nips/XuRLJ14}). Gramm loss \cite{DBLP:conf/cvpr/GatysEB16} can satisfy the two mentioned conditions, however, unlike CL, it does not constrain the content of the generated image as it describes the image globally.

The CL function can be calculated between the source and the generated images, or between the target (ground-truth) and the generated images. The source-generated loss aims at saving the details of the source image such as detailed boundaries. The target-generated loss maintains the properties of the target image in the generated image, e.g. target image style. In our case, as will be presented in the next section, the source and target training image pairs are of identical faces. Therefore, the target-generated loss also maintains the detailed properties of the object in the source image.

Both losses were calculated between image embeddings extracted by a pre-trained VGG19 \cite{DBLP:journals/corr/SimonyanZ14a} network trained on the ImageNet database \cite{imagenet_cvpr09}. The total loss is calculated as given in \cite{DBLP:conf/eccv/MechrezTZ18} and formulated as:
\begin{equation} 
\small
\begin{split}
L_{CX}(s,t,g,l_s,l_t)= & \lambda_1( -log(CX(\Phi^{l_s}_1(g),\Phi^{l_s}(s))) ) + \\ & \lambda_2  (-log(CX(\Phi^{l_t}_2(g),\Phi^{l_t}(t)))),
\end{split}
\end{equation}
where $s$, $t$, and $g$ are the are the source, target, and generated images respectively.
$CX$ is the rotation and scale invariant contextual similarity \cite{DBLP:conf/eccv/MechrezTZ18}.
$\Phi$ is a perceptual network, VGG19 in our work. $\Phi^{l_s}(x)$, $\Phi^{l_t}(x)$ are the embeddings vectors extracted from the image $x$ at layer $l_s$ and $l_t$ of the perceptual network respectively. Here $l_s$ is the \texttt{conv4\char`_2} and $l_t$ is the \texttt{conv3\char`_2} and \texttt{conv4\char`_2} layers, as motivated in \cite{DBLP:conf/eccv/MechrezTZ18}. In our implementation, $\lambda_1 = 0.01$ and $\lambda_2 = 0.99$ by checking the resulting generated image visually. Moreover, as mentioned earlier, because the pairs of source and target images are of identical faces, the loss weighted by $\lambda_2$ maintains the structural details of the source image. An illustration of the proposed CRN-based synthesis approach is shown in Fig. \ref{fig:CRN}. 

\begin{figure*}[h]
                \centering
\includegraphics[width=1.5\columnwidth]{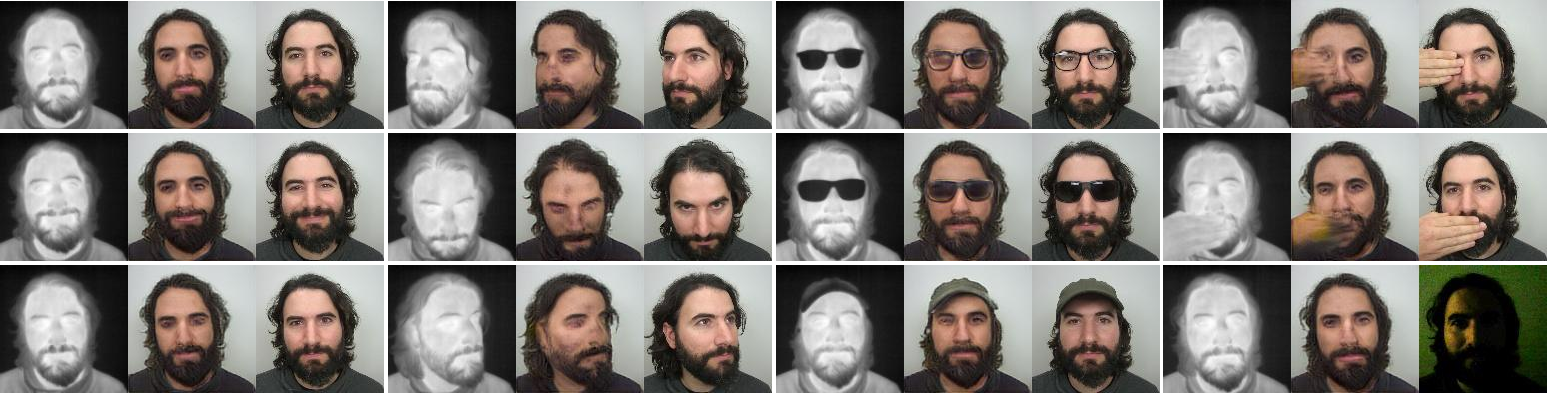}
\vspace{-3mm}
\caption{Samples of visible images generated from thermal captures. Each group of three images have the source thermal image (O-THM) on the left, the generated visible image (G-VIS) in the middle and the ground-truth (O-VIS) on the right.}
\label{img:samples}
\vspace{-6mm}
\end{figure*} 

\textbf{Baseline approaches:}
(1) PIX2PIX: the approach proposed by Isola et al. \cite{DBLP:conf/cvpr/IsolaZZE17} learns the mapping from one domain to another, by training a Conditional GAN using Least Absolute Deviations (L1) loss function. The generator is based on U-Net \cite{DBLP:conf/miccai/RonnebergerFB15} architecture, an encoder-decoder with skip connections between mirrored layers in the encoder and decoder stacks. The discriminator aims to classify real images from generated ones.
The training was run for 85 epochs, batch size of one, and 2e-4 learning rate.

(2) TV-GAN: the approach proposed by Zhang et al. \cite{TZhang18} builds a network to generate visible-like face images from thermal captures. This work is inspired by PIX2PIX, as it uses the same network for the generationr. However, the authors proposed a multi-task discriminator, that doesn't only classify real from generated images, but also performs a closed-set face recognition to obtain identity loss. This aims to generate visible-like images while preserving identity information from the thermal inputs. The training was run for 65 epochs, batch size of one, and 2e-4 learning rate.

\vspace{-3mm}

\section{Experimental setup}
\label{sec:exp}
\vspace{-2mm}



To develop and evaluate our solution, we use the VIS-TH face publicly available database \cite{DBLP:conf/biosig/Mallat18}. The database contains face images in visible spectrum with pixel resolution of 1920$\times$1080 and their thermal counterpart of pixel resolution of 160$\times$120. Face images from both spectra were acquired simultaneously using the dual sensor camera Flir DUO R \cite{flir} with a spectral response range of 7.5 - 13.5$\mu$m for the thermal sensor. 50 subjects of different ages, gender, and ethnicities have participated in the data collection. Unlike the few existing databases of visible and thermal face, this database includes a wide range of variations: expression, illumination, pose and occlusion. Considering 21 facial variation for each subject, the database contains in total 2100 images.


Thermal images were up-sampled from 120x120 to 128x128 pixels to match the network input. The visible images were down-sampled to 128x128 pixels as well. Visible images are treated as three color channel image by the network, while the thermal image (one channel) is fed identically to all three channels. To enable a realistic evaluation of our approach, and because the face alignment solutions in the thermal spectrum is not a well standardized technique, the face images were not aligned. Therefore, the data contained images with slightly shifted faces that also did not always have the same scale in comparison to the image size.



The solution introduced in the previous section is trained and deployed to generate visible images from thermal ones. 
The generation is trained always on the ground-truth thermal and visible images (containing all capture variations, except the completely dark visible images) of 45 (out of 50) identities and the resulted solution is used to generate the images for the remaining 5 identities in the database. This was performed 10 times to test our generation on all the images in the database without overlapping the test and train images or identities. We trained the solution on a system with system i5-6500 CPU 16 GB RAM and NVIDIA GTX 1050 Ti 4 GB on-board. The training was run for 40 epochs, patch size of one, and 1e-4 learning rate. The training takes slightly under 3 hours.

In the next section, we present and discuss a visual qualitative samples of the generated visible images. To have a quantitative comparison of the generated image quality, we present a number of image quality measures (suggested in \cite{wasnik2017assessing}) for the original and generated images. Namely, as defined in the relative references, the sharpness \cite{gao2007standardization}, blur \cite{narvekar2011no}, exposure \cite{shirvaikar2004optimal}, global contrast factor (GCF) \cite{matkovic2005global}, contrast \cite{zohra2017linear, gao2007standardization}, light symmetric (LS) \cite{wasnik2017assessing}, and brightness \cite{zohra2017linear}. The same experimental details and quality metrics were used to train and evaluate the baseline PIX2PIX \cite{DBLP:conf/cvpr/IsolaZZE17} and TV-GAN \cite{TZhang18}.

\begin{figure}[h]
                \centering
\begin{subfigure}[b]{0.160\columnwidth}
                \centering
\includegraphics[width=0.95\columnwidth]{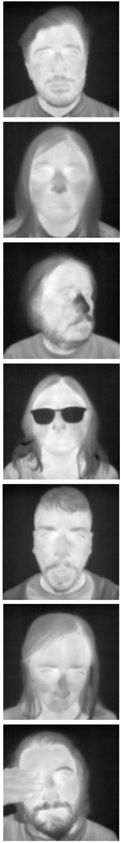}
\caption{}
\label{fig:sc1}
\end{subfigure}
\begin{subfigure}[b]{0.160\columnwidth}
                \centering
\includegraphics[width=0.95\columnwidth]{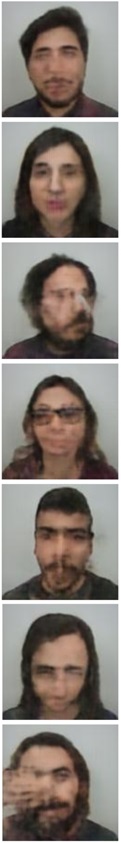}
\caption{}
\label{fig:sc2}
\end{subfigure}
\begin{subfigure}[b]{0.160\columnwidth}
                \centering
\includegraphics[width=0.95\columnwidth]{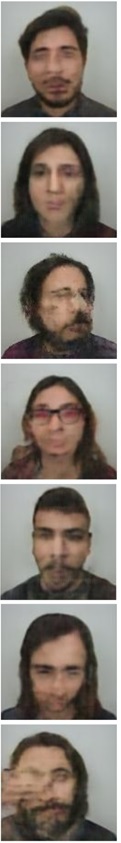}
\caption{}
\label{fig:sc3}
\end{subfigure}
\begin{subfigure}[b]{0.160\columnwidth}
                \centering
\includegraphics[width=0.95\columnwidth]{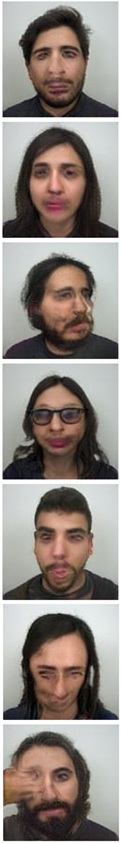}
\caption{}
\label{fig:sc4}
\end{subfigure}
\begin{subfigure}[b]{0.160\columnwidth}
                \centering
\includegraphics[width=0.95\columnwidth]{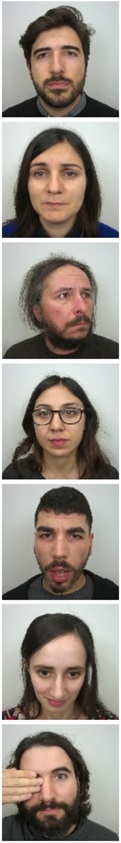}
\caption{}
\label{fig:sc5}
\end{subfigure}
\vspace{-3mm}
\caption{Visual comparison between our approach and two recently proposed solutions under different scenarios. (a) thermal, (b) PIX2PIX \cite{DBLP:conf/cvpr/IsolaZZE17}, (c) TV-GAN\cite{TZhang18}, (d) ours, (e) g.-truth. }
\label{img:samples_comp}
\vspace{-5mm}
\end{figure}

\vspace{-4.0mm}

\section{Results}
\label{sec:res}
\vspace{-2.0mm}


\begin{figure}[h]
                \centering
\includegraphics[width=0.80\columnwidth]{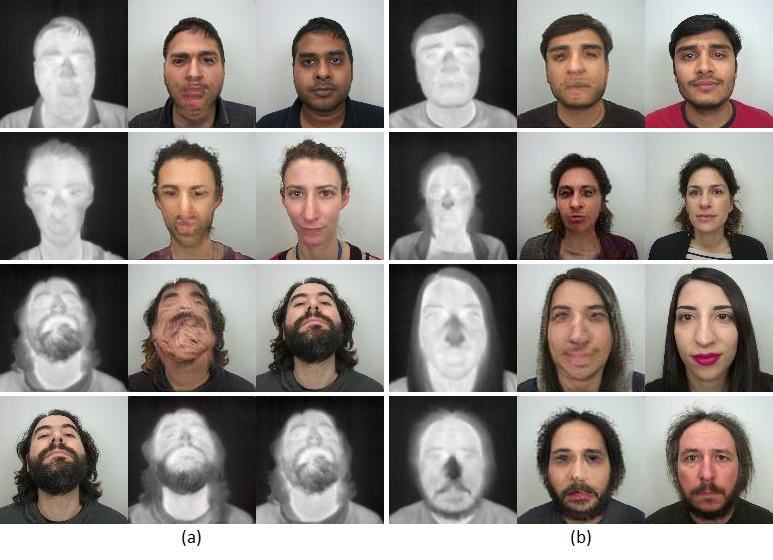}
\vspace{-3.5mm}
\caption{Samples of generated images under challenging scenarios (a) and of different genders, scale, and ethnicities (b). }
\label{img:lessThanPerfect}
\vspace{-6.5mm}
\end{figure}

\def\arraystretch{1.0}
\begin{table*}[h]
\scriptsize
\setlength\tabcolsep{0.15cm}
  \begin{center}
  \begin{tabular}{l | c | c | c | c | c | c | c }
      \toprule 
      \textbf{} & \textbf{Sharpness}  & \textbf{Blur}  & \textbf{Exposure} & \textbf{GCF}  & \textbf{Contrast}  & \textbf{LS} & \textbf{Brightness} \\
      \midrule 
         \textbf{O-VIS} & 0.215 ($\pm0.030$)  & 0.520 ($\pm0.071$) & 0.208 ($\pm0.039$) & 8.851 ($\pm1.886$) & 0.472 ($\pm0.031$) & 0.056 ($\pm0.041$) &0.589 ($\pm0.062$)\\
				         \textbf{O-THM} & 0.171 ($\pm0.016$)  & 0.281 ($\pm0.068$) & 0.286 ($\pm0.024$) & 11.565 ($\pm1.927$) & 0.311 ($\pm0.020$) & 0.085 ($\pm0.053$) &0.398 ($\pm0.037$)\\
				         \textbf{TV-GAN \cite{TZhang18}} & 0.161 ($\pm0.016$)  & 0.368 ($\pm0.139$) & 0.179 ($\pm0.0158$) & 7.788 ($\pm0.906$) & 0.492 ($\pm0.0095$) & 0.037 ($\pm0.022$) &0.533 ($\pm0.022$)\\
								\textbf{PIX2PIX \cite{DBLP:conf/cvpr/IsolaZZE17}} & 0.162 ($\pm0.020$)  & 0.361 ($\pm0.136$) & 0.179 ($\pm0.0169$) & 7.783 ($\pm0.960$) & 0.491 ($\pm0.0091$) & 0.037 ($\pm0.021$) &0.535 ($\pm0.028$)\\
								         \textbf{G-VIS (ours)} & \textbf{0.213 ($\pm0.023$)}  & \textbf{0.555 ($\pm0.056$)} & \textbf{0.226 ($\pm0.0243$)} & \textbf{9.590 ($\pm1.074$)} & \textbf{0.487 ($\pm0.013$)} & \textbf{0.044 ($\pm0.027$)} &\textbf{0.582 ($\pm0.034$)}\\
				\midrule
    \end{tabular}
		\vspace{-2.8mm}
    \caption{Quality measures of the original ground-truth thermal (O-THM) and visible (O-VIS) captures, as well as our generated visible-like (G-VIS) images. Each of the measures is given as a mean value over the full database ($\pm$ standard deviation). The quality similarity between the visible generated and ground-truth data is noted (most similar to O-VIS in bold).}
    \label{tab:quality}
  \end{center}
\vspace{-7.5mm}
\end{table*}


The images in Fig. \ref{img:samples} are samples of the generated visible images from thermal captures. For each sample, we show the source image (O-THM), the generated visible image (G-VIS), and the ground-truth visible image (O-VIS), in this order from left to right. The first two columns of samples show variations in facial expressions and pose. The third column shows samples with occlusion by clear eyeglasses, sun glasses, and head-wear. The fourth column includes occlusions by hand, as well as non-optimal illumination in the visible capture. The different frontal pose generated images contains minimum large scale artifact, however, small details (e.g. around the eye) are less than optimal. Different poses still produced visually realistic images, with slightly more artifacts compared to the frontal poses. This might be due to containing larger areas (e.g. around the nose) where the thermal information is relatively similar. Artificial occlusions were generated in relative good quality, as they commonly block much of the thermal signal. Some confusion was noticed between regenerating transparent and sun glasses, as they both have similar thermal print. Generating visible images with occlusion by hand was successful, however, with high level of blur in the hand region. As expected, the amount of the visible illumination does not affect the generated visible image from the thermal one, which is one of the main motivation of this work. Figure \ref{img:samples_comp} presents a visual comparison, in different scenarios, between images generated by our proposed solution and the baseline TV-GAN \cite{TZhang18} and PIX2PIX \cite{DBLP:conf/cvpr/IsolaZZE17} solutions. One can notice in the figure that our solution produces a more detailed generation of the visible-like images.

Fig. \ref{img:lessThanPerfect}.a shows some of the few challenging situations that produced less optimal visible results. The first (top) visible image generation sample produced a visible image that differs in the color tone from the ground-truth image, although darker than the images presented in Fig. \ref{img:samples}. This is mainly due to the fact that different skin tones appears close to identical in the thermal domain. To a less degree, this might be also caused by the limited ethnic variability of the training data. The second sample from the top of Fig. \ref{img:lessThanPerfect}.a shows another visually distorted generated sample. Here, the generation seems to add a facial hair-like distortion when compared to the ground-truth image. This can be reasoned by the limited female gender representation within the training data, and the limited size of the training data in general. The last two samples of the Fig. \ref{img:lessThanPerfect}.a presents the image generation, in both directions, for an extreme pose. In this case, the visible image generated from the thermal image seems to face some confusion in the chin area. However, in the bottom sample, this is not problematic when generating a thermal-like image from the visible image. Fig. \ref{img:lessThanPerfect}.b presents samples of generated visible images from different genders, ethnicities, and face scale.


For a quantitative comparison between the generated and original ground-truth images, Table \ref{tab:quality} presents image quality measures calculated from these datasets. The overall similarity between the generated and ground-truth data is clear. In general, the generated images are slightly more blur and less sharp. The exposure is slightly increased when generating a visible image
. The contrast and GCF is higher in the generated visible image when compared to the ground-truth.
The generated data scored lower LS values in comparison to the corresponding ground-truth, which indicate a higher lightning symmetry in the generated images.
Finally, the brightness of the generated data did not differ significantly from the corresponding ground-truth.
Overall, the generated images maintained, to a large degree, the quality of the corresponding ground-truth images. This similarity in quality metrics to the ground-truth is held to a significantly lower degree by the two baseline solutions, TV-GAN \cite{TZhang18} and PIX2PIX \cite{DBLP:conf/cvpr/IsolaZZE17}, also presented in Table \ref{tab:quality}.

\vspace{-5mm}

\section{Conclusion}
\label{sec:con}
\vspace{-2.5mm}

Driven by the suboptimal deployment scenarios of face recognition, and motivated by the need for cross-spectrum manual and automatic recognition, we present a solution for high-quality color visible image generation from thermal captures. Unlike previous works, our solution does not require large training database, data augmentation, polarimetric sensors, high computational needs, or low generated image resolution. The proposed solution is based on cascaded refinement networks and contextual loss, making it relatively easy to train on small databases and inherently scale and rotation invariant. The results proved, by comparison to two state-of-the-art solutions, the visual high quality of the generated images under different scenarios, as well as, the quantitative quality similarity to the ground-truth images.

\vspace{-3mm}

\bibliographystyle{IEEEbib}
\bibliography{strings,refs}

\begin{thebibliography}{10}

\bibitem{HZhang17}
H.~Zhang, V.~M. Patel, B.~S. Riggan, and S.~Hu,
\newblock ``Generative adversarial network-based synthesis of visible faces
  from polarimetrie thermal faces,''
\newblock in {\em 2017 IEEE International Joint Conference on Biometrics
  (IJCB)}, Oct 2017, pp. 100--107.

\bibitem{Riggan18}
B.~S. Riggan, N.~J. Short, and S.~Hu,
\newblock ``Thermal to visible synthesis of face images using multiple
  regions,''
\newblock in {\em 2018 IEEE Winter Conference on Applications of Computer
  Vision (WACV)}, March 2018, pp. 30--38.

\bibitem{TZhang18}
T.~Zhang, A.~Wiliem, S.~Yang, and B.~Lovell,
\newblock ``{TV-GAN}: Generative adversarial network based thermal to visible
  face recognition,''
\newblock in {\em 2018 International Conference on Biometrics (ICB)}, Feb 2018,
  pp. 174--181.

\bibitem{PLS15}
Shuowen Hu, Jonghyun Choi, Alex~L. Chan, and William~Robson Schwartz,
\newblock ``Thermal-to-visible face recognition using partial least squares,''
\newblock {\em J. Opt. Soc. Am. A}, vol. 32, no. 3, pp. 431--442, Mar 2015.

\bibitem{Sarfraz15}
R.~Stiefelhagen M.~Sarfraz,
\newblock ``Deep perceptual mapping for thermal to visible face recognition,''
\newblock {\em International Journal of Computer Vision}, pp. 1--11, 2015.

\bibitem{Li07}
Jun Li, Pengwei Hao, Chao Zhang, and Mingsong Dou,
\newblock ``Hallucinating faces from thermal infrared images,''
\newblock in {\em Proceedings of the International Conference on Image
  Processing, {ICIP} 2008, October 12-15, 2008, San Diego, California, {USA}}.
  2008, pp. 465--468, {IEEE}.

\bibitem{DBLP:conf/iccv/ChenK17}
Qifeng Chen and Vladlen Koltun,
\newblock ``Photographic image synthesis with cascaded refinement networks,''
\newblock in {\em {IEEE} International Conference on Computer Vision, {ICCV}
  2017, Venice, Italy, October 22-29, 2017}. 2017, pp. 1520--1529, {IEEE}
  Computer Society.

\bibitem{DBLP:conf/eccv/MechrezTZ18}
Roey Mechrez, Itamar Talmi, and Lihi Zelnik{-}Manor,
\newblock ``The contextual loss for image transformation with non-aligned
  data,''
\newblock in {\em {ECCV} 2018, Munich, Germany, September 8-14, 2018,
  Proceedings}. 2018, vol. 11218, pp. 800--815, Springer.

\bibitem{DBLP:conf/cvpr/IsolaZZE17}
Phillip Isola, Jun{-}Yan Zhu, Tinghui Zhou, and Alexei~A. Efros,
\newblock ``Image-to-image translation with conditional adversarial networks,''
\newblock in {\em {CVPR} 2017, Honolulu, HI, USA, July 21-26, 2017}. 2017, pp.
  5967--5976, {IEEE}.

\bibitem{DBLP:conf/nips/XuRLJ14}
Li~Xu, Jimmy S.~J. Ren, Ce~Liu, and Jiaya Jia,
\newblock ``Deep convolutional neural network for image deconvolution,''
\newblock in {\em Annual Conference on Neural Information Processing Systems
  2014, December 8-13 2014, Montreal, Quebec, Canada}, 2014, pp. 1790--1798.

\bibitem{DBLP:conf/cvpr/GatysEB16}
Leon~A. Gatys, Alexander~S. Ecker, and Matthias Bethge,
\newblock ``Image style transfer using convolutional neural networks,''
\newblock in {\em {CVPR} 2016, Las Vegas, NV, USA, June 27-30, 2016}. 2016, pp.
  2414--2423, {IEEE}.

\bibitem{DBLP:journals/corr/SimonyanZ14a}
Karen Simonyan and Andrew Zisserman,
\newblock ``Very deep convolutional networks for large-scale image
  recognition,''
\newblock {\em CoRR}, vol. abs/1409.1556, 2014.

\bibitem{imagenet_cvpr09}
J.~Deng, W.~Dong, R.~Socher, L.-J. Li, K.~Li, and L.~Fei-Fei,
\newblock ``{ImageNet: A Large-Scale Hierarchical Image Database},''
\newblock in {\em CVPR09}, 2009.

\bibitem{DBLP:conf/miccai/RonnebergerFB15}
Olaf Ronneberger, Philipp Fischer, and Thomas Brox,
\newblock ``U-net: Convolutional networks for biomedical image segmentation,''
\newblock in {\em {MICCAI} 2015, Munich, Germany, October 5 - 9, 2015,
  Proceedings, Part {III}}. 2015, vol. 9351, pp. 234--241, Springer.

\bibitem{DBLP:conf/biosig/Mallat18}
{K}hawla {M}allat and {J}ean-{L}uc {D}ugelay,
\newblock ``A benchmark database of visible and thermal paired face images
  across multiple variations,''
\newblock in {\em International Conference of the Biometrics Special Interest
  Group, {BIOSIG} 2018, Darmstadt, Germany, September}. 2018, {LNI}, {GI} /
  {IEEE}.

\bibitem{flir}
``Flir systems,'' http://www.flir.com/.

\bibitem{wasnik2017assessing}
Pankaj Wasnik, Kiran~B Raja, Raghavendra Ramachandra, and Christoph Busch,
\newblock ``Assessing face image quality for smartphone based face recognition
  system,''
\newblock in {\em Biometrics and Forensics (IWBF), 2017 5th International
  Workshop on}. IEEE, 2017, pp. 1--6.

\bibitem{gao2007standardization}
Xiufeng Gao, Stan~Z Li, Rong Liu, and Peiren Zhang,
\newblock ``Standardization of face image sample quality,''
\newblock in {\em International Conference on Biometrics}. Springer, 2007, pp.
  242--251.

\bibitem{narvekar2011no}
Niranjan~D Narvekar and Lina~J Karam,
\newblock ``A no-reference image blur metric based on the cumulative
  probability of blur detection (cpbd),''
\newblock {\em IEEE Transactions on Image Processing}, vol. 20, no. 9, pp.
  2678--2683, 2011.

\bibitem{shirvaikar2004optimal}
Mukul~V Shirvaikar,
\newblock ``An optimal measure for camera focus and exposure,''
\newblock in {\em System Theory, 2004. Proceedings of the Thirty-Sixth
  Southeastern Symposium on}. IEEE, 2004, pp. 472--475.

\bibitem{matkovic2005global}
Kresimir Matkovic, L{\'a}szl{\'o} Neumann, Attila Neumann, Thomas Psik, and
  Werner Purgathofer,
\newblock ``Global contrast factor-a new approach to image contrast.,''
\newblock {\em Computational Aesthetics}, vol. 2005, pp. 159--168, 2005.

\bibitem{zohra2017linear}
Fatema~Tuz Zohra, Andrei~Dmitri Gavrilov, Omar~Zatarain Duran, and Marina
  Gavrilova,
\newblock ``A linear regression model for estimating facial image quality,''
\newblock in {\em Cognitive Informatics \& Cognitive Computing (ICCI* CC), 2017
  IEEE 16th International Conference on}. IEEE, 2017, pp. 130--138.

\end{thebibliography}

\end{document}